\documentclass[journal]{IEEEtai}

\usepackage[colorlinks,urlcolor=blue,linkcolor=blue,citecolor=blue]{hyperref}
\usepackage{color,array}
\usepackage{graphicx}
\usepackage{subfigure}
\usepackage{booktabs}
\usepackage{amsmath,amssymb}
\usepackage[utf8]{inputenc}

\begin{document}

\title{Understanding LLM Scientific Reasoning through Promptings and Model’s Explanation on the Answers}

\author{Alice Rueda\textsuperscript{*}, \IEEEmembership{Member, IEEE}, Mohammed S. Hassan\textsuperscript{*}, Argyrios Perivolaris\textsuperscript{*}, Bazen G. Teferra, Reza Samavi, Sirisha Rambhatla, Yuqi Wu, Yanbo Zhang, Bo Cao, Divya Sharma, Sridhar Krishnan\textsuperscript{\textdagger}, \IEEEmembership{Member, IEEE}, and Venkat Bhat\textsuperscript{\textdagger}

\thanks{\textsuperscript{*}co-first authors, \textsuperscript{\textdagger}co-senior authors}

\thanks{VB is supported by an Academic Scholar Award from the University of Toronto Department of Psychiatry and has received research funding from the Canadian Institutes of Health Research, Brain \& Behavior Foundation, Ontario Ministry of Health Innovation Funds, Royal College of Physicians and Surgeons of Canada, Department of National Defence (Government of Canada), New Frontiers in Research Fund, Associated Medical Services Inc. Healthcare, American Foundation for Suicide Prevention, Roche Canada, Novartis, and Eisai.}

\thanks{ A. Rueda is with the Interventional Psychiatry Program, St. Michael’s Hospital, Unity Health Toronto, Toronto, Ontario, Canada (e-mail: alice.rueda@unityhealth.to).}

\thanks{M. S. Hassan is with the Department of Electrical, Computer, and Biomedical Engineering, Toronto Metropolitan University, Toronto(e-mail: mohammed.siadhassan@torontomu.ca).}

\thanks{A. Perivolaris is with the Interventional Psychiatry Program, St. Michael’s Hospital, Unity Health Toronto, Toronto, Ontario, Canada (e-mail: argyrios.perivolaris@unityhealth.to).}

\thanks{B.G. Teferra is with the Interventional Psychiatry Program, St. Michael’s Hospital, Unity Health Toronto, Toronto, Ontario, Canada (e-mail: bazengashaw.teferra@unityhealth.to).}

\thanks{R. Samavi is with the Department of Electrical, Computer, and Biomedical Engineering, Toronto Metropolitan University, Toronto(e-mail: samavi@torontomu.ca).}

\thanks{S. Rambhatla is with the Department of Management Science and Engineering, University of Waterloo, Ontario, Canada (e-mail: sirisha.rambhatla@uwaterloo.ca).}

\thanks{Y. Wu is with the Department of Electrical and Computer Engineering, University of Alberta, Canada (e-mail:yuqi14@ualberta.ca).}

\thanks{Y. Zhang is with the Department of Psychiatry, Faculty of Medicine and Dentistry, University of Alberta, Edmonton, Alberta, Canada (e-mail: yanbo9@ualberta.ca).}

\thanks{B. Cao is with the Department of Psychiatry, Faculty of Medicine and Dentistry, University of Alberta, Edmonton, Alberta, Canada (e-mail: cloudbocao@gmail.com).}

\thanks{D. Sharma is with the Department of Mathematics and Statistics, York University, Ontario, Canada (e-mail:divya03@yorku.ca).}

\thanks{S. Krishnan is with the Department of Electrical, Computer, and Biomedical Engineering, Toronto Metropolitan University, Toronto(e-mail: krishnan@torontomu.ca).}

\thanks{V. Bhat is with the Interventional Psychiatry Program, St. Michael’s Hospital, Unity Health Toronto, Toronto, Ontario, Canada and the Department of Psychiatry, University of Toronto, Toronto, Ontario, Canada (e-mail: venkat.bhat@utoronto.ca).}}


\maketitle

\begin{abstract}
Large language models (LLMs) have demonstrated remarkable capabilities in natural language understanding, reasoning, and problem-solving across various domains. However, their ability to perform complex, multi-step reasoning tasks—essential for applications in science, medicine, and law—remains an area of active investigation. This paper examines the reasoning capabilities of contemporary LLMs, analyzing their strengths, limitations, and potential for improvement. The study uses prompt engineering techniques on the Graduate-Level GoogleProof Q\&A (GPQA) dataset to assess the scientific reasoning of GPT-4o. Five popular prompt engineering techniques and two tailored promptings were tested: baseline direct answer (zero-shot), chain-of-thought (CoT), zero-shot CoT, self-ask, self-consistency, decomposition, and multipath promptings. Our findings indicate that while LLMs exhibit emergent reasoning abilities, they often rely on pattern recognition rather than true logical inference, leading to inconsistencies in complex problem-solving. The results indicated that self-consistency outperformed the other prompt engineering technique with an accuracy of 52.99\%, followed by direct answer (52.23\%). Zero-shot CoT (50\%) outperformed multipath (48.44\%), decomposition (47.77\%), self-ask (46.88\%), and CoT (43.75\%). Self-consistency performed the second worst in explaining the answers. Simple techniques such as direct answer, CoT, and zero-shot CoT have the best scientific reasoning.  We propose a research agenda aimed at bridging these gaps by integrating structured reasoning frameworks, hybrid AI approaches, and human-in-the-loop methodologies. By critically evaluating the reasoning mechanisms of LLMs, this paper contributes to the ongoing discourse on the future of artificial general intelligence and the development of more robust, trustworthy AI systems
\end{abstract}

\begin{IEEEImpStatement}
The development of large language models (LLMs) has streamlined tasks such as language translation, summarization, and writing. LLMs outperform humans in general knowledge tasks even with zero-shot promptings. This study uses prompt engineering techniques on the Graduate-Level GoogleProof Q\&A (GPQA) dataset to assess the scientific reasoning of GPT-4o.  Five popular prompt engineering techniques and two tailored promptings were tested: baseline direct answer (zero-shot), chain-of-thought (CoT), zero-shot CoT, self-ask, self-consistency, decomposition, and multipath promptings. The results indicated that self-consistency outperformed the other prompt engineering technique with an accuracy of 52.99\%, followed by direct answer (52.23\%). Zero-shot CoT (50\%) outperformed multipath (48.44\%), decomposition (47.77\%), self-ask (46.88\%), and CoT (43.75\%). Self-consistency performed the second worst in explaining the answers. Simple techniques such as direct answer, CoT, and zero-shot CoT have the best scientific reasoning. This study establishes a baseline for evaluating GPT-4o’s scientific reasoning and optimizing prompt engineering strategies.
\end{IEEEImpStatement}


\section{Introduction}

\IEEEPARstart{L}{arge} Language Models (LLMs) are largely trained with general knowledge and can be used directly for knowledge retrieval purposes, such as in zero-shot prompting or with one or more examples in few-shot prompting. Despite LLMs being demonstrated to mimic cognitive processes of humans \cite{niu2024large}, particularly in language processes \cite{niu2024large}, \cite{zhang2024mulcogbench}, their reasoning capability is still limited. The current architecture and training paradigm require models to be equipped with specific content knowledge and optimized prompting to reduce hallucination and improve performance. Often, research will focus on how to improve task performance through prompt engineering and less on tracking the reasoning process of an LLM. This paper uses GPT-4o as an example to investigate the scientific reasoning capability of LLMs with no access to additional information or external tools. This paper aims to illustrate the scientific reasoning capability of LLMs using GPT-4o on the Graduate-Level Google-Proof Q\&A (GPQA) dataset as an example by analyzing the responses and explanations to various prompt engineering techniques. The GPQA is a challenging dataset of 448 multiple-choice questions written by content experts in biology, physics, and chemistry.

\section{Background}

Despite the growing body of research on LLMs and their performance across various tasks, most studies have primarily focused on improving accuracy through prompt engineering rather than investigating how these models reason. Prior work has largely evaluated LLMs on task completion, such as mathematical problem-solving \cite{ahn2024large} or common-sense reasoning \cite{chang2024explaining}. However, these studies typically measure performance improvements without systematically analyzing the underlying reasoning process that leads to a model’s answers. The GPQA dataset \cite{rein2023gpqa}, while previously used to test the accuracy of LLMs on scientific questions, has not yet been applied to investigate the reasoning strategies employed by these models. Prior to DeepSeekR1 \cite{deepseek2025r1}, accuracy, rather than explanation, has been the de-facto performance measure for scientific reasoning capability (see Appendix A for frontier LLM performance). A few methods have been proposed to improve accuracy on GPQA dataset, including more sophisticated prompt engineering techniques such as Iteration of Thought \cite{radha2024iteration}, Story of Thought \cite{javadi2024can}, Multi-thinking Modes Tree \cite{li2024mtmt}, make use of analogies \cite{yuan2024boosting}, or using multiple LLMs \cite{omar2024refining}. Assessing both accuracy and reasoning quality across different prompting techniques can provide new insights into the multifaceted interplay of an LLM’s processes and justification for complex scientific problems, bridging a critical gap in the literature.

Prompt engineering is a technique used to improve instructions to obtain desirable and more structured responses from LLMs that are pre-trained in general tasks. Using well-crafted prompts, LLMs excel in context-aware retrieval tasks, understanding user intent, and can even use tools to perform more complex tasks through API calls. However, LLMs can sometimes present false or misleading information as facts, often referred to as hallucinations. Some well-known techniques reduce hallucination, such as Retrieval Augmented Generation (RAG) \cite{bechard2024reducing} and ReACT \cite{yao2022react} using content to provide grounding. Others, like Chain-of-Thought (CoT)\cite{wei2022chain} and self-consistency \cite{wang2022self}, are designed for reasoning and logic. CoT-like prompts constitute the majority of the reasoning and logic prompts \cite{sahoo2024systematic}, including Auto-CoT \cite{zhang2022automatic}, LogiCoT \cite{liu2023logicot}, Chain-of-Symbol \cite{hu2023chain}, Tree of Thought \cite{yao2023tree}, Thread of Thought \cite{zhou2023thread}, and Chain of Table prompting \cite{wang2024chain}. 

Prompt engineering has been used to perform reasoning tasks \cite{qiao2022reasoning}, especially in mathematical reasoning where CoT, self-consistency, and Program-of-Thought (separating reasoning steps from final computation) were used on various models to solve mathematical problems \cite{ahn2024large}. The CoT’s superiority was demonstrated on 23 of BIG-Bench’s reasoning tasks. \cite{suzgun2022challenging} Chen et al. \cite{chen2023mcc} proposed Multi-CoT Consistent Knowledge Distillation (MCC-KD) to enforce consistency with the questions.  This teacher-student-based CoT technique enforces consistency by using multiple teacher LLMs to extract rationales on the problem, N-gram, and Jaccard similarity to filter similar rationales to maintain diversity. The smaller language models (students) learn to answer the question correctly by measuring similarity through bi-directional KLD on their rationales with the teachers’ rationales. The most common approach to engineering a prompt is to tailor the prompt to tasks and datasets. 

Besides mathematical reasoning using GSM8K, there are other cognitive problems LLMs are tasked with that require other reasoning processes such as Commonsense QA datasets and AI2's Reasoning Challenge (ARC), general reasoning BIG-Bench \cite{srivastava2022beyond}, understanding and generating code HumanEval \cite{chen2021evaluating}, and scientific reasoning dataset GPQA. Benchmark datasets for natural language processing are widely available and well defined, such as CoLA, GLUE, and Benchmark of Linguistic Minimal Pairs (BLiMP). \cite{warstadt2020blimp} Cognitive benchmarking datasets are limited to a few well-known ones, such as CogBench \cite{coda2024cogbench}, and MulCogBench2. CogBench consists of seven cognitive psychology experiments (tasks) to measure 10 behavioral metrics. These tasks include probabilistic reasoning, two-armed bandit tasks (horizon and restless bandit tasks), instrumental learning, two-step tasks, temporal discounting, and balloon analog risk tasks. The 10 metrics are priori weighting, likelihood weighting, directed exploration, random exploration, meta-cognition, learning rate, optimism bias, model-basedness, temporal discounting, and risk-taking. MulCogBench is a multimodal dataset that enables researchers to relate LLM cognitive processes to human brain activity. The cognitive dataset contains subjective semantic ratings, eye-tracking, functional magnetic resonance imaging (fMRI), and magnetoencephalography (MEG). However, a more domain-specific dataset, such as GPQA has a lower data leakage risk, which provides a measure of the reasoning process. GPQA is designed for scalable oversight problems. 

The scalable oversight problem is when LLMs are superior in domain knowledge yet allow humans to supervise, evaluate, and guide to solve the problems. \cite{bowman2022measuring} GPQA consists of extremely difficult scientific multiple-choice questions with four choices, with domain experts achieving 65\% accuracy. At the same time, the solutions provided in the dataset also come with an explanation. The explanation provides another measure to judge and guide the answers. GPQA is designed to evaluate a model’s reasoning capability on top of content retrieval. As titled in the paper, GPQA is a graduate-level Google-proof Q\&A benchmark dataset across physics, chemistry, and biology subdomains in quantum mechanics, astrophysics, organic chemistry, genetics, and molecular biology.  These multiple-choice questions are high-quality, extremely difficult, and their content is not easily accessible. Non-domain experts achieved 34\% accuracy with access to the internet, and 60-80\% by domain experts who have completed or are currently pursuing their doctoral degree. The GPQA has been used to develop better prompts for benchmark model performance. To the best of our knowledge, GPQA has not been used to explain the scientific reasoning of LLMs.    

By prompting the model to answer the GPQA questions and to provide an explanation of the answer, we can measure not just the accuracy of the model’s problem-solving ability but also understand the reasons behind such an answer. GPQA provides a set of explanations that can be used as ground truth to compare a model's explanation. This paper uses basic promptings on GPT-4o to study LLM’s scientific reasoning. The model’s knowledge has been restricted to Dec 2023 to ensure no data leakage and model checkpoint on April 1, 2024, to establish reproducibility. The contributions of this study are 1) providing better insight into the scientific reasoning behind LLMs, 2) proposing an objective approach to measure reasoning using GPQA, and 3) providing an example to demonstrate automated scalable oversight problems.

\section{Methods}

The experiments were conducted using GPT-4o-2024-08-06 as the candidate model and lowered the temperature setting to 0.5 to reduce randomness. Lower temperatures, closer to 0.0, tend to produce more deterministic responses, whereas higher temperatures promote exploration and variability. To understand the model's reasoning ability, we limited the knowledge cutoff set to December 2023 to ensure that no data leakage accidentally occurred through HFRL or continuous content releases and set the model date to April 1, 2024. The multiple-choice questions in GPQA were presented one after another with four possible choices shuffled. In order to capture all explanations, the token limit was set to 4,096 for prompt completion.

\subsection{GPQA Dataset}

This GPQA dataset is designed to benchmark the advancement in the reasoning capability of LLMs as the models evolve, achieving this through the concept of scalable oversight methods. Scalable oversight allows skilled non-experts (i.e., completed or pursuing doctoral degrees in other domains) to verify LLM’s reasoning steps. One of the key components of scalable oversight is interpretability, where LLMs provide logical explanations throughout the process. With an accuracy of 78\%, GPT-o1 has demonstrated its performance in part with human experts [5].  This study uses all 448 questions provided in the GPQA-main dataset to ensure all questions are difficult and objective. In the supplementary material, Table D1 in Appendix D provides a breakdown of each subdomain in the GPQA-main dataset with their accuracy answered by experts and non-experts.

\subsection{Prompt Engineering Techniques}

Prompt engineering techniques are generally versatile and sometimes can benefit from engineering a combined method tailored to the tasks. This versatility allowed us to observe LLM’s response to scientific reasoning. We implemented prompts that were classified as reasoning and logic \cite{sahoo2024systematic} in some that were not mentioned but demonstrated the ability to elicit an LLM’s reasoning capability, and two of our interpretations and creations. We explored the behaviour of the model using seven prompt engineering techniques: direct answer (zero-shot), chain-of-thought (CoT), zero-shot CoT, decomposition - an adapted implementation of decomposed prompting with its decomposed prompting’s reasoning chains, modular problem-solving and compositionality, self-ask, self-consistency, and our proposed multipath prompting. Each of these techniques is summarized below:

\begin{enumerate}
\item[{\it 1)}]{\it Direct Answer (Zero-Shot)}

This was a baseline reference for other techniques. The model was asked to solve a multiple-choice problem.

\item[{\it 2)}]{\it Chain-of-thought (Cot)}

This explicit reasoning approach was guided by three examples to show the model’s step-by-step reasoning to solve multiple-choice reasoning problems.

\item[{\it 3)}]{\it Zero-Shot CoT}

This prompting asked the model to solve the multiple-choice problem by "Let's think step by step" before each answer without providing any examples. 

\item[{\it 4)}]{\it Self-Ask}

For complex questions, intermediate steps or information are required to reduce the gaps. Self-ask allows the models to decompose questions effectively and generate correct intermediate answers. This technique was designed to handle multi-hop problems by asking the LLMs to check if there are intermediate steps by asking relevant follow-up questions and answering them before reaching the final answer.

\item[{\it 5)}]{\it Self-Consistency}

Self-consistency replaces the “greedy decode” with a diverse set of reasoning paths and marginalizes the reasoning paths to aggregate the final answers that are the most consistent. As the temperature setting was reduced to 0.5 to reduce randomness, the number of paths defined for self-consistency allowed a small number. We chose the minimum setting of a small number of three to enable selection. The use of three samples allows increased control and decreased randomness associated with the technique.

\item[{\it 6)}]{\it Decomposition}

This technique asked the LLM to break down complex problems recursively into simple manageable subproblems. All subproblems were sent to the model to solve in a step-by-step manner. The final results integrated solutions to all subproblems. We tried two different ways of implementing subproblem prompting. Option 1 was to send subproblems sequentially, one at a time, until all subproblems receive completion before composition. Option 2 concatenated all subproblems and sent everything at once.

\item[{\it 7)}]{\it Multipath Prompting}

This prompt was created to observe the model's behaviour using a mixture of logical and illogical steps. This prompt asked the model to create a justification or verification for each of the four choices. Then, based on their justification to answer whether the choice is correct or not. A final prompt was sent with all the justifications and answers to ask the LLM to decide which one was correct.

\end{enumerate}

\subsection{Performance Measures}

Besides measuring accuracy, we have captured which questions were answered correctly or incorrectly by all prompts and recorded all details to examine the reasoning capability of the model. After answering each question, GPT-4o was also asked to provide an explanation for its answer. GPQA provided an explanation feature in the dataset, providing the explanation to reach the solution. Cosine similarity was used to measure the similarity between the ground truth explanations from GPQA and the explanations elicited from the prompts. Cosine similarity ($S_c$) has known problems for underestimating frequent words with other same words across different contexts (within words) or different words in similar context (across words). \cite{zhou2022problems} The underestimating effect has no impact on the explanations for GPQA dataset as the contents are highly specialized. A pre-trained MPNet \cite{song2020mpnet} sentence transformer model was used to detect sentence semantic similarity in the explanation of prompts and ground truth. Then, we look at the effect distance of the prompts on explanations elicited using stand mean distance on the explanation supported by Jaccard distance on the final answers, i.e. 1- Jaccard similarity.

\section{Results}

\subsection{Accuracy of Answers}

After testing all prompt engineering techniques, Option 1 generated unreliable answers due to hallucination and had to be removed from the analysis. Except for self-consistency and decomposition-option 2, the number of correct answers was summarized for different prompt engineering techniques. 

The accuracy of each prompt, represented as a percentage out of 100\%, is provided in Table 1. Table 1 provides a summary of the accuracy of the questions, correct answers to the three domains, corresponding accuracy over the 448 questions, and the number of questions with missing answers. The techniques with the most correct answers ranked in descending order are self-consistency (237/448, 52.99\%), direct answer (234/448, 52.23\%), zero-shot CoT (224/448, 50\%), multipath (217/448, 48.44\%), decomposition (214/448, 47.77\%), and CoT (196/448, 43.75\%). A breakdown of correctness for each subdomain can be found in Supplementary Tables B1 and B2 in Appendix B. Some prompts failed to provide answers at all. GPT-4o did not provide an answer to questions when its perceived answer was not listed as one of the multiple choices. This reasoning process is captured in the returned explanation. Except for direct answer, all other prompt engineering techniques disagreed with all the choices available in one or more questions. The number and questions with no answers are listed under Not Answered in Table 1. Six of the seven prompts did not provide answers to question 257. Only a direct answer has provided both the answer and explanation, despite the answer being incorrect. All other prompts elicited disagreement.

\begin{table*}
    \centering
    \caption{Performance Metrics for Different Reasoning Methods}
        \begin{tabular}{lccccccc}
            \toprule
            & Direct Answer & ZS CoT & CoT & Decomposition & Self-Ask & Self-Consistency & MultiPath \\
            \midrule
            Correct & 234 & 224 & 196 & 214 & 210 & 237 & 217 \\
            Jaccard similarity & 0.235 & 0.267 & 0.200 & 0.19 & 0.211 & 0.364 & 0.2 \\
            Chemistry & 106 & 96 & 86 & 91 & 89 & 105 & 90 \\
            Biology & 38 & 35 & 30 & 37 & 36 & 38 & 31 \\
            Physics & 90 & 93 & 80 & 86 & 85 & 94 & 96 \\
            Accuracy & 52.23\% & 50.00\% & 43.75\% & 47.77\% & 46.88\% & 52.90\% & 48.44\% \\
            \midrule
            Not Answered \\
            Missing Count & 0 & 1 & 1 & 2 & 1 & 2 & 4 \\
            Missing Questions & - & \#257 & \#257 & \#19, \#257 & \#257 & \#78, \#257 & \#19, \#78, \#257, \#286 \\
            \bottomrule
        \end{tabular}
    \label{tab:performance}
\end{table*}

Fig. 1 illustrates the cumulative correctness throughout the GPQA-main dataset in (a) and modelled with linear regression in (b). The slope represents the mean, and the distance from the slope represents the variation. A higher R-squared value demonstrates that the mean can be an accurate representation of accuracy. All prompts have a linear characteristic on the number of correct answers, as supported by the linear models presented in the Supplementary Table C1 in Appendix C. The coefficients (slopes) of the linear models are very close to the accuracy values, as all R-squared scored over 0.99. The direct answer has the steepest slope of 0.539, followed by self-consistency’s 0.528. As expected, CoT has the shallowest slope of 0.438.

\begin{figure*}[htp]
  \centering
  \subfigure[Accumulated Correctness]{\includegraphics[width=0.47\linewidth]{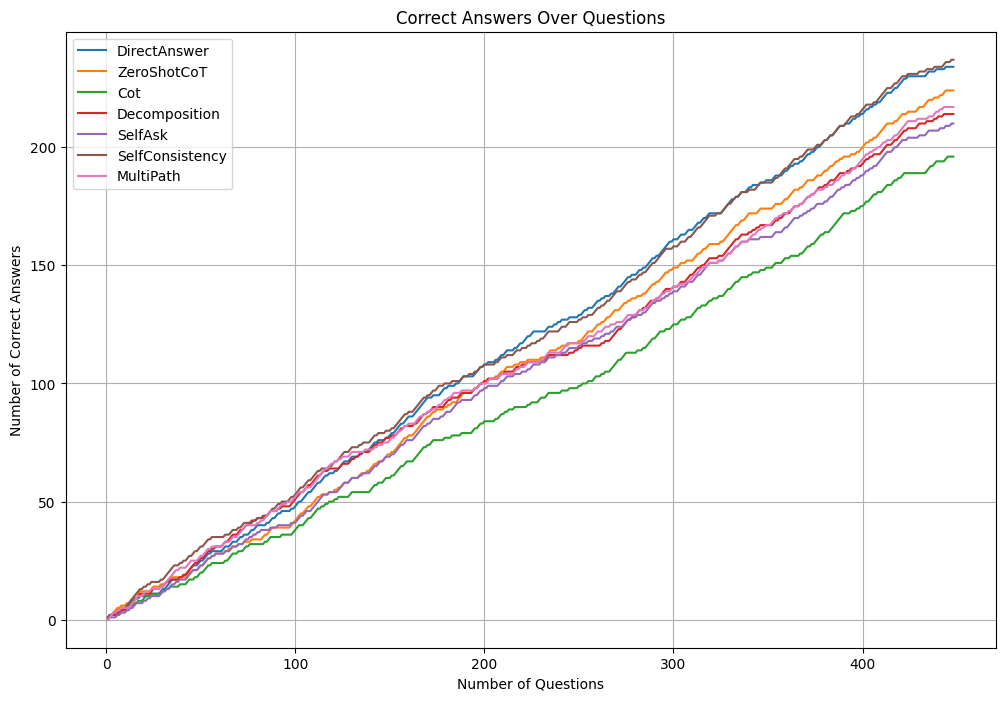}}\quad
  \subfigure[Linear approximation]{\includegraphics[width=0.47\linewidth]{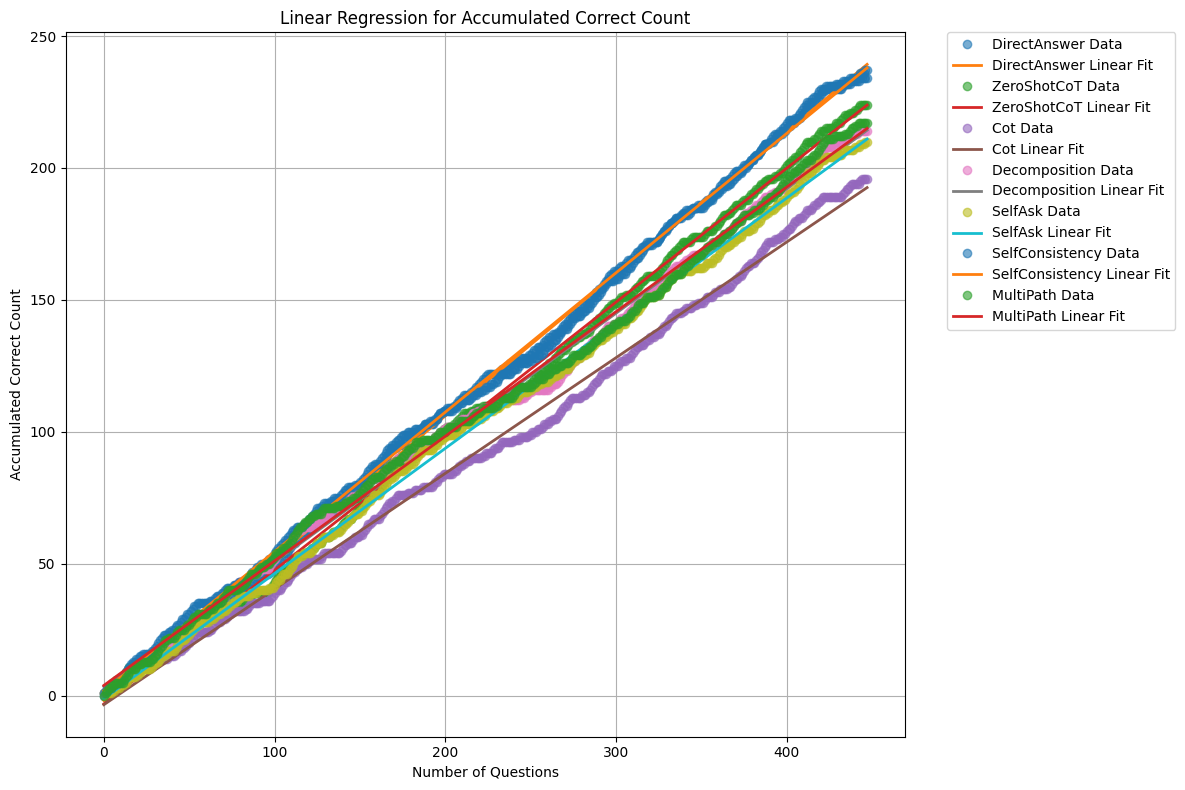}}
  \caption{Cumulated correctness. (a) Each correct answer for the prompts is stored as a cumulative sum. (b) Model the accumulated correctness with linear regression.}
  \label{fig:1}
\end{figure*}

As the GPQA-main dataset has the easiest questions removed, this reduces the chance of all prompts answering the questions correctly. The majority of the questions were answered correctly by some prompts and incorrectly by others. There is a significant percentage of questions that were answered unanimously correct or incorrect. To better understand the occurrence of correctness, Fig. 2 indicates the locations of the questions that were answered correctly by all prompts and incorrectly by all prompts. There are 98/448 (21.86\%) questions that were answered correctly by all, 104/448 (23.21\%) were answered incorrectly by all, and the rest 246 (54.91\%) were answered correctly by some. Fig. 2a presents the cumulated correct answers after sorting by subdomain and plotting on top of various coloured subdomains. To locate which questions were unanimously answered correctly and incorrectly, Fig. 2b added a vertical line to indicate the questions that were correctly answered (in green) and incorrectly answered (in red) by all prompts. Fig. 3a and 3b are also matched in regions with correctness. The dense red or green vertical lined regions are reflected by the shallowness and steepness of the slopes.

\begin{figure*}[htp]
  \centering
  \subfigure[Correctness over subdomains]{\includegraphics[width=0.47\linewidth]{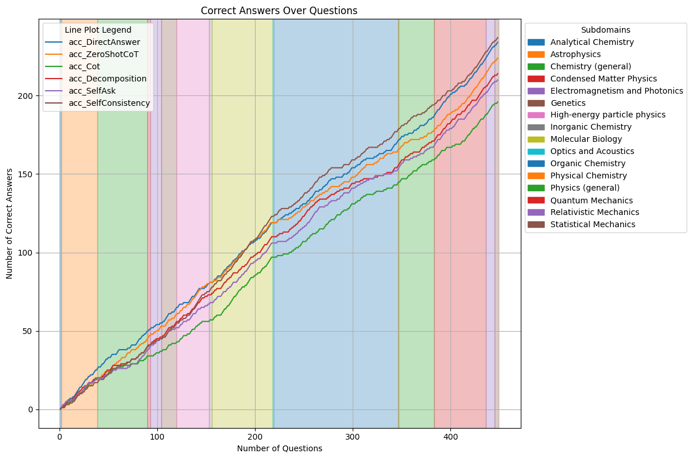}}\quad
  \subfigure[All correct or wrong]{\includegraphics[width=0.47\linewidth]{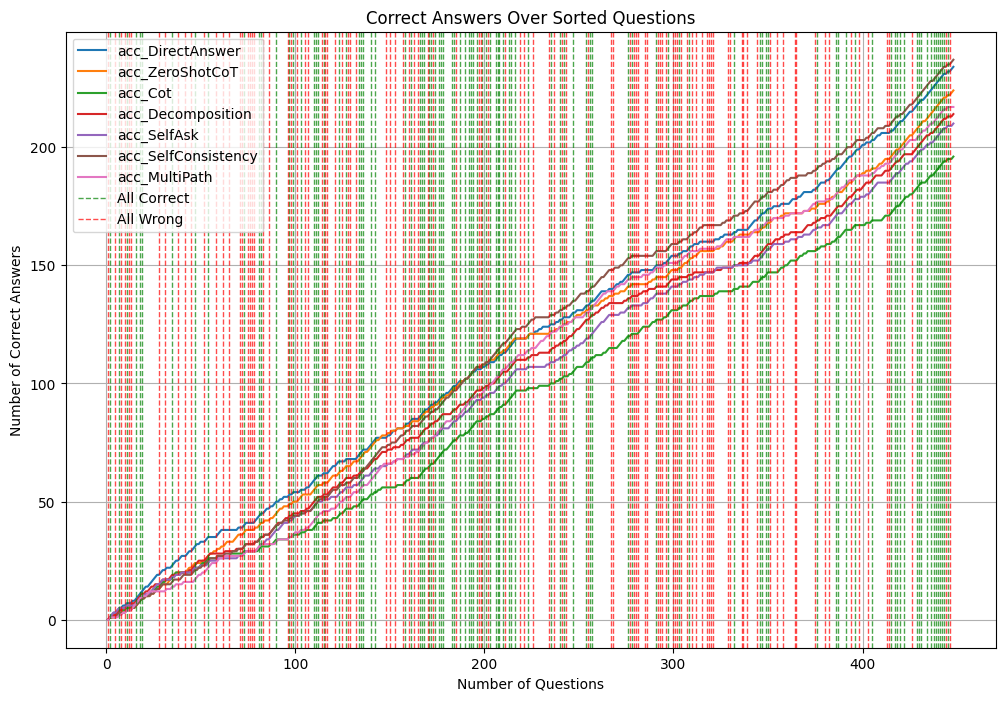}}
  \caption{Cumulated correctness sorted by subdomain. The answers were first sorted by subdomain before calculating the accumulated sum. (a) Cumulated correctness after sorting is plotted on top of the subdomains. Overlaying the cumulated sum on the subdomain question type. (b) Visual indication on the questions that were answered correctly (in green) and incorrectly (in red) by all prompts.}
  \label{fig:2}
\end{figure*}

\subsection{Scientific Reasoning}

Explanation quality varied significantly across techniques. With each answer, GPT-4o was asked to provide justification (explanation) for its choice of answer. This explanation provides insights into its scientific reasoning process. Appendix C presents samples of the explanation and answers for each prompt. The prompt-induced explanation was used to compare with GPQA’s ground truth explanation. Table 2 summarizes the cosine similarity measure on the explanations elicited from the prompts. The distribution of the similarity measures is not Gaussian distributed as indicated by the Shapiro-Wilk’s test where all p-values were less than 0.05. Ranking the similarity by a median, the explanation elicited by the prompts in descending order is direct answer, CoT, zero-shot CoT, self-ask, decomposition, multipath, and self-consistency. Using psychology’s definition of correlation \cite{akoglu2018correlation}, Table 2 also groups the cosine similarity into weak (both positive and negative), moderate, and strong similarities between the prompt-induced explanation and ground truth. The overarching explanations provided by simple prompts, such as direct answer, zero-shot CoT, CoT, and self-ask have strong similarities to the ground truth. Multipath prompting (self-consistency and multipath) and subproblem-based decomposition achieved moderate similarity. Fig. 3 illustrates the histogram of the similarities using a bin size of 0.02 overlaid with kernel density estimation. 

\begin{table*}
    \centering
    \caption{Cosine similarity ($S_c$) of GPT-4o’s explanation to the ground truth on GPQA-main}
        \begin{tabular}{lccccccc}
            \toprule
            & Direct Answer & ZS CoT & CoT & Decomposition & Self-Ask & Self-Consistency & MultiPath \\
            \midrule
            Shapiro-Wilk (W) & 0.915 & 0.954 & 0.934 & 0.963 & 0.822 & 0.844 & 0.814 \\
            \midrule
            Statistics \\
            Median (IQR) & 0.733 (0.136) & 0.730 (0.140) & 0.736 (0.137) & 0.694 (0.144) & 0.727 (0.133) & 0.645 (0.200) & 0.684 (0.406) \\
            Mean (SD) & 0.718 (0.114) & 0.730 (0.140) & 0.726 (0.111) & 0.678 (0.110) & 0.705 (0.144) & 0.587 (0.215) & 0.559 (0.288) \\
            Minimum & 0.031 & 0.168 & 0.164 & 0.299 & -0.043 & -0.065 & -0.076 \\
            Maximum & 0.916 & 0.936 & 0.915 & 0.889 & 0.908 & 0.883 & 0.900 \\
            \midrule
            Similarity Strength \\
            Weak: $S_c<0.3$ (n) & 2 & 1 & 3 & 1 & 11 & 55 & 110 \\
            Moderate: $0.3\leq S_c < 0.7$ (n) & 164 & 171 & 158 & 234 & 170 & 242 & 136 \\
            Strong: $S_c \geq 0.7$ (n) & 282 & 276 & 287 & 213 & 267 & 151 & 202 \\
            \midrule
            \textbf{Overall Similarity Strength} & Strong & Strong & Strong & Moderate & Strong & Moderate & Moderate \\
            \bottomrule
        \end{tabular}
    \label{tab:cosine_similarity}
\end{table*}

\begin{figure}
\centerline{\includegraphics[width=18.5pc]{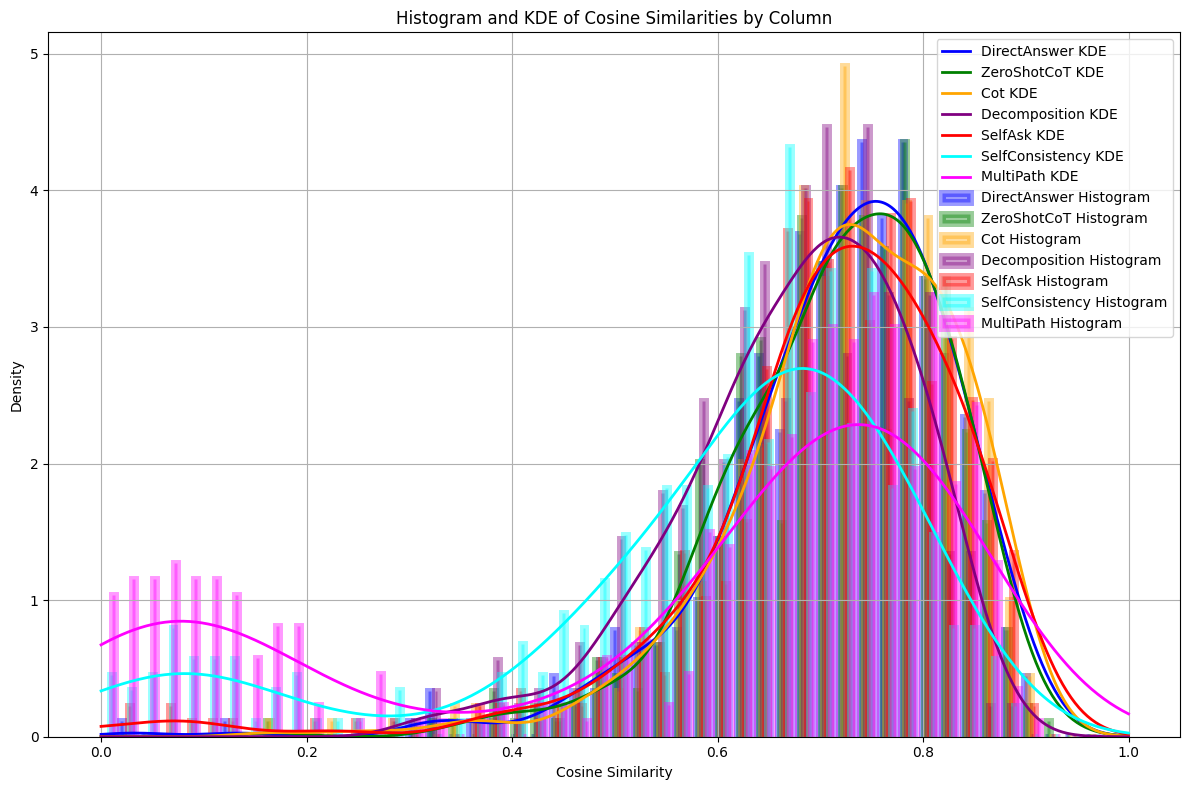}}
\caption{Distribution of the cosine similarity that GPT-4o’s explanation for each prompt to the ground truth explanation provided in GPQA-main dataset.}
\end{figure}

The accuracy rankings reflect how well each prompt can reason through scientific questions, with some techniques outperforming others. The direct answer, CoT, and zero-shot CoT provided the closest explanation to the ground truth with very similar explanation measures. While self-consistency scored the highest in accuracy, its explanation measure was ranked last. Self-ask, multipath, and self-consistency all have the median inside the moderate similarity category. CoT and direct answer are very similar in their explanations compared to the ground truth. The heatmap in Fig. 4 shows where the model’s explanations were very different from the ground truth. The dissimilarity of self-consistency and multipath visibly stands out with the close to 0 and negative values.

\begin{figure}
\centerline{\includegraphics[width=18.5pc]{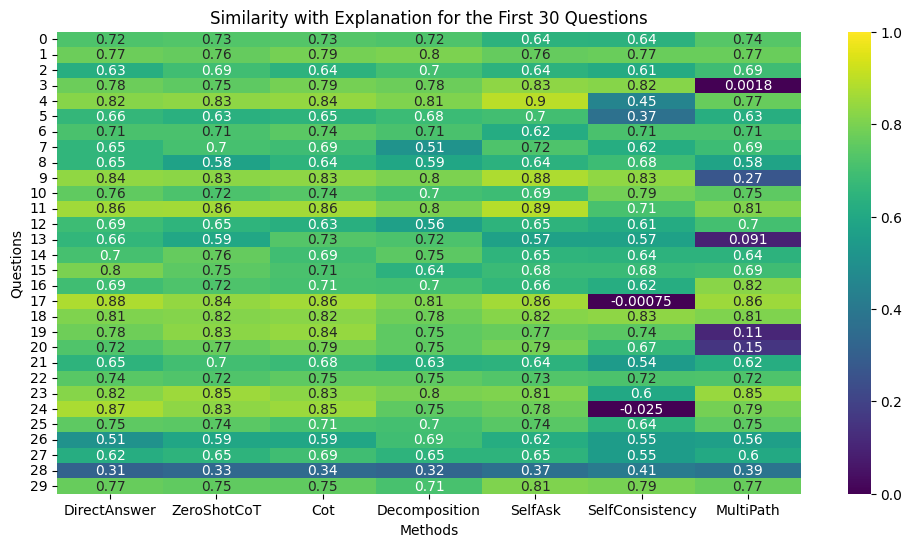}}
\caption{Cosine similarity for the first 30 questions that GPT-4o’s explanation for each prompt to the ground truth explanation provided in GPQA-main dataset.}
\end{figure}

\section{Discussion}

This study evaluated the scientific reasoning capabilities of GPT-4o using the GPQA-main dataset, emphasizing the interplay between accuracy and explanation quality across various prompt engineering techniques. The findings illuminate critical insights into the model's reasoning abilities and the impact of prompt strategies on performance. GPT-4o with its knowledge set to December 2023 and model checkpoint on April 1, 2024, was used to study scientific reasoning ability using single path (direct answer, zero-shot CoT, CoT, decomposition, and self-ask) and multipath prompt engineering techniques. The model temperature setting was lowered to 0.5 to minimize hallucinations while still permitting variability in responses, particularly for self-consistency (3 paths) and the in-house designed multipath. The maximum number of completion token was set to 4,096 as required to explain GPQA questions \cite{muennighoff2025s1} and to avoid mistakes caused by overthinking.\cite{chen2024overthinking} The multipath technique was intentionally designed to try to induce hallucination by justifying for all the provided choices before answering the questions. 

Self-consistency and direct answer led to the performance in accuracy with self-consistency slightly higher in the number of correct answers and direct answer having a slightly steeper linear regression slope. CoT performed the worst in accuracy. Much like DeepSeekR1 that “may be hurt” by a few-shot CoT, GPT-4o performed better with zero-shot CoT to CoT. 

With high R-squared scores ($R2>0.99$), the slope of the cumulative correctness reflected the accuracy of all prompt techniques. The correctness for sorted subdomains matched the steepness of the slope to the vertical green (all prompts answered correctly) and red (all prompts answered incorrectly) lines. GPT-4o refuses to answer some of the questions, especially question 257. The explanation provided by GPT-4o provided a glance at the model’s reasoning process to reach or not reach the answer. Compared to the ground truth explanation provided by the dataset, CoT achieved the highest cosine similarity, in descending order, followed by the direct answer, zero-shot CoT, self-ask, decomposition, multipath, and self-consistency. Although self-consistency demonstrated its superiority in answering the questions correctly, it performed poorly in providing or not providing explanations, as illustrated in Fig. 4. On the other hand, the explanations provided by multipath were as expected, as GPT-4o was forced to provide justifications for the wrong answers.    

When subproblems were presented sequentially, hallucinations increased. To avoid this, we suggest concatenating subproblems and presenting them all at once, which helps reduce hallucination. By concatenating subproblems and sending them all at once, hallucination was managed. Surprisingly, the multipath prompting did not elicit the expected hallucination as GPT-4o was asked to provide justification for incorrect answers in the multiple-choice questions. 

On the other hand, GPT-4o could refuse to answer if the provided choices did not match its explanation for all prompts, except for the direct answer prompt. In question 257, the model noted that none of the choices were correct but selected the closest one due to the constraints of the prompt. The instruction for direct answer was simply asking GPT-4o to answer the question, thus GPT-4o was forced to answer the question even though its answer was not listed in any of the choices as provided in the explanation provided for questions 257, “None of these options match \( 9.6 \times 10^{6} \) GeV.”  This indicates GPT-4o is able to prioritize tasks as demonstrated by the explanation, “However, since I am required to choose an option, and acknowledging the possibility of a typographical error or miscommunication in the provided data, we will select the closest option, though it is not correct based on the calculation.” All other prompts reached the same calculation but refused to choose an answer. Despite this result, the accuracy of GPT-4o’s responses is consistent, as shown by the high R-squared scores, which indicate that the model’s accuracy does not fluctuate wildly. The slope of the accuracy curve serves as a reliable indicator of performance in answering complex scientific questions.

The areas with dense vertical green lines (indicating correct answers from all prompts) correspond to subdomains where GPT-4o performed well. In contrast, regions with dense red lines (incorrect answers from all prompts) suggest subdomains where the model struggled. This allowed us to infer that GPT-4o is well-trained in certain subdomains, and vice versa in the shallow slope subdomains where the region is populated with dense vertical red lines. 

GPT-4o is less confused in its reasoning when given direct or simple prompts as demonstrated by CoT, direct answer, and zero-shot CoT. Decomposition elicited the best explanations when the problem was first broken down into smaller manageable subproblems. The strong similarity in GPT-4o’s explanation to the ground truth supports its scientific reasoning ability to answer non-multiple-choice questions except for prompt engineering techniques that cannot elicit reasonable explanations. For some questions, self-consistency only provided the answers with no explanation. Single-path prompts are more straightforward for the model to reason about than multipath prompts. With the multipath technique, the explanations for each path can be too diverse and yet consistent for the model to choose the explanation. As there is only one temperature setting for path creation and answer selection, probabilistic reasoning is common for both the path creation and final decision-making. If the model accepts two levels of temperature setting, one for exploration to create paths and another for final decision making, the result might be improved from exploiting the final answer. 

This study has a few limitations to consider. Our studies are conducted by non-experts with no doctoral degrees in chemistry, physics, or biology. The study can benefit from domain experts’ grading of the explanations elicited by the prompts beyond the cosine similarity measures. The study can also benefit from capturing the explanation for all the paths in addition to the current final answer’s explanation. This will require increasing the token size limit and will drive the cost up. However, the current recorded responses provide large explanatory data that can be used for deeper analysis in the future. These findings can serve as a baseline for future studies on GPT-4o’s reasoning capabilities and act as a baseline to compare with other models.

\section{Conclusions}

In conclusion, this study provides an evaluation of GPT-4o’s scientific reasoning capabilities using the GPQA dataset, highlighting the interplay between accuracy and explanation quality across different prompt engineering techniques. While self-consistency achieved the highest accuracy, direct answer, CoT, and zero-shot CoT generated explanations more similar to the ground truth, suggesting that reasoning quality does not always correlate with accuracy. Our findings underscore the need for future research to refine prompt strategies that optimize both correctness and interpretability. This work serves as a foundation for further studies on LLM reasoning, particularly in domains requiring high levels of interpretability and trust.

\section{Appendices}

\subsection{GPQA Diamond Performance Comparison on Frontier Models (Table III)}

\begin{table*}
    \centering
    \caption{Best-performed setting for each of the well-known reasoning model family on GPQA Diamond dataset}
        \begin{tabular}{lcccccc}
            \toprule
            & OpenAI-o1-0912 & DeepSeekR1-Zero & Claude3.5 Sonnet & Llama3 405B & Qwen2.5-72B & ChatGLM-4 \\
            Accuracy & 77.3\% & 73.3\% & 59.4\% & 51.1\% & 45.9\% & 39.3\% \\
            \bottomrule
        \end{tabular}
    \label{tab:other_models}
\end{table*}

\subsection{Correctness Counted by Subdomain (Table IV)}

\begin{table*}
    \centering
    \caption{Count of correctly answered questions in each subdomain}
        \begin{tabular}{lccccccc}
            \toprule
            & Direct Answer & ZS CoT & CoT & Decomposition & Self-Ask & Self-Consistency & MultiPath \\
            \midrule
            Analytical Chemistry (1) &	0	1 &	0 &	1 &	0 &	1 &	1 \\
            Astrophysics (37) &	22 & 19	& 24 & 21 & 21 &	21 & 23 \\
            Chemistry (general) (51) & 28 &	22 & 21 & 24 & 22 &	22 & 20 \\
            Condensed Matter Physics (3) &	2 &	2 &	2 &	2 &	2 &	3 &	2 \\
            Electromagnetism and Photonics (11) &	4 &	4 &	2 &	7 &	2 &	6 &	6 \\
            Genetics (16) &	8 &	5 &	4 &	5 & 6 &	6 &	7 \\
            High-energy particle physics (33) &	18 & 20	& 16 & 15 &	17 &	18 & 17 \\
            Inorganic Chemistry (3) & 2 &	2 &	1 &	1 &	2 &	2 &	1 \\
            Molecular Biology (62) & 30 &	30 & 26 & 32 & 30 & 32 & 24 \\
            Optics and Acoustics (1) & 1 &	1 &	1 &	0 &	1 &	1 &	1 \\
            Organic Chemistry (127) & 76 &	71 & 64	& 65 & 65 & 80 & 68 \\
            Physical Chemistry (1) & 0 & 0 &	0 & 0 &	0 &	0 &	0 \\
            Physics (general) (36) & 16 &	15 & 12 & 15 & 16 & 17 & 18 \\
            Quantum Mechanics (53) & 22 &	25 & 20 & 23 & 21 & 22 & 24 \\
            Relativistic Mechanics (9) & 4 &	5 &	2 &	2 &	3 &	4 &	4 \\
            Statistical Mechanics (4) & 1 &	2 &	1 &	1 &	2 &	2 &	1 \\
            \midrule
            Total &	234 & 224 &	196 & 214 & 210 & 237 &	217 \\
            \bottomrule
        \end{tabular}
    \label{tab:subdomain_performance}
\end{table*}

\subsection{Linear Regression Table (Table V)}

\begin{table*}
    \centering
    \caption{Linear regression to model the correctness of the seven prompts}
        \begin{tabular}{lccccccc}
            \toprule
            & Direct Answer & ZS CoT & CoT & Decomposition & Self-Ask & Self-Consistency & MultiPath \\
            \midrule
            Coefficient (Slope) & 0.5392 &	0.5077 & 0.4383 & 0.4717 &	0.4745 & 0.5281 & 0.4722 \\
            Intercept & -1.7721 & -3.1078 & -3.3615 & 3.4674 & -1.1140 & 1.8873 & 3.9226 \\
            R\textsuperscript{2} Score & 0.9992 & 0.9984 & 0.9965 & 0.9967 & 0.9988 & 0.9982 & 0.9969 \\
            \bottomrule
        \end{tabular}
    \label{tab:linear_regression}
\end{table*}

\subsection{GPQA Questions and Human Performance (Table VI)}

\begin{table*}
    \centering
    \caption{Subdomain questions in the GPQA-main dataset and the corresponding human performance}
        \begin{tabular}{lcccc}
            \toprule
            Domain & Questions & Expert Accuracy & Non-Expert Accuracy & Expertise Gap \\
            \midrule
            Molecular Biology & 62 & 75.8 & 37.1 & 38.7 \\
            Physics (general) & 36 & 63.9 & 29.6 & 34.3 \\
            Organic Chemistry & 127 & 86.6 & 27.8 & 58.8 \\
            Chemistry (general) & 51 & 58.8 & 28.1 & 30.7 \\
            Relativistic Mechanics & 9 & 77.8 & 18.5 & 59.3 \\
            Quantum Mechanics & 53 & 62.3 & 30.2 & 32.1 \\
            Electromagnetism and Photonics & 11 & 72.7 & 24.2 & 48.5 \\
            Genetics & 16 & 75.0 & 41.7 & 33.3 \\
            High-energy particle physics & 33 & 60.6 & 30.3 & 30.3 \\
            Astrophysics & 37 & 62.2 & 31.5 & 30.6 \\
            Statistical Mechanics & 4 & 75.0 & 33.3 & 41.7 \\
            Inorganic Chemistry & 3 & 100.0 & 33.3 & 66.7 \\
            Condensed Matter Physics & 3 & 33.3 & 33.3 & 0.0 \\
            Physical Chemistry & 1 & 100.0 & 66.7 & 33.3 \\
            Optics and Acoustics & 1 & 100.0 & 0.0 & 100.0 \\
            Analytical Chemistry & 1 & 0.0 & 0.0 & 0.0 \\
            \bottomrule
        \end{tabular}
    \label{tab:GPQA_dataset}
\end{table*}

\bibliographystyle{IEEEtran} 
\bibliography{references}

\section{Acknowledgment}

None.

\end{document}